\documentclass[conference]{IEEEtran}
\IEEEoverridecommandlockouts
\usepackage{cite}
\usepackage{amsmath,amssymb,amsfonts}
\usepackage{algorithmic}
\usepackage{algorithm}
\usepackage{graphicx}
\usepackage{textcomp}
\usepackage{xcolor}
\def\BibTeX{{\rm B\kern-.05em{\sc i\kern-.025em b}\kern-.08em
    T\kern-.1667em\lower.7ex\hbox{E}\kern-.125emX}}
\begin{document}

\title{MCTS Based Agents for Multistage Single-Player Card Game}

\author{
\IEEEauthorblockN{Konrad Godlewski}
\IEEEauthorblockA{\textit{Warsaw University of Technology}  \\ Warsaw, Poland \\ jowisz11@gmail.com}
\and
\IEEEauthorblockN{Bartosz Sawicki}
\IEEEauthorblockA{\textit{Warsaw University of Technology} \\ Warsaw, Poland \\ bartosz.sawicki@pw.edu.pl}
}

\maketitle

\begin{abstract}
The article presents the use of Monte Carlo Tree Search algorithms for the card game Lord of the Rings. The main challenge was the complexity of the game mechanics, in which each round consists of 5 decision stages and 2 random stages. To test various decision-making algorithms, a game simulator has been implemented. The research covered an agent based on expert rules, using flat Monte-Carlo search, as well as complete MCTS-UCB. Moreover different playout strategies has been compared. As a result of experiments, an optimal (assuming a limited time) combination of algorithms were formulated. The developed MCTS based method have demonstrated a advantage over agent with expert knowledge.
\end{abstract}


\section{Introduction}

\textit{The Lord of the Rings: The card game} is one of the most popular card games. Since its launch in 2011 by Fantasy Flight Games, it has gained great popularity, as evidenced by more than 100 official extensions, dozens of blogs and millions of fans around the world. The uniqueness and enormous success of this game is due to its cooperative character and the fact that it can only be played by one person, but the core set supports up to 2 players. The players have to fight against the random pile of cards that represents the forces of Sauron, which are obstacles to be overcome.

The Monte-Carlo tree search (MCTS) is a stochastic algorithm. It proved its unique power in 2016 by beating human master in Go game, which has been described as a last moment when human players had a chance to compete with AI players. Since that time there is growing number of applications of MCTS in various games~\cite{b2},\cite{b5},\cite{b7}. Even popular card game games such as Magic: The Gathering, has been studied in terms of MCTS~\cite{Ward2009}.

However applications in cooperative and single-player games~\cite{Turkay2012} are still quite limited. In this paper MCTS algorithm is successfully used in non-competitive game, which is challenging player with high levels of complexity and randomness. 

The MCTS algorithm has been created as a tool for perfect information games (such as Go or Chess). However it could be treated as a general purpose computational intelligence algorithm. There are several successful efforts in applications in different scientific domains such as logistic optimization~\cite{Powley2012} and chemistry synthesis planning~\cite{Segler2018}.



\section{The Lord of the Rings: The Card Game}

"The Lord of the Rings: The Card Game", often abbreviated as LoTR, is a complicated cooperative game with several decision-making stages. Understanding its mechanics is necessary when searching the optimal strategy for the player. Especially that the element of the project was to implement a game simulator.

Before starting the game, two decks of cards \textit{player deck} and \textit{encounter deck} are shuffled. The player sets his initial threat level, which is the sum of the appropriate parameters of all three heroes, then draws six cards into his hand from the \textit{player's deck}. 

\begin{figure}[htbp]
    \centerline{\includegraphics[width=140pt]{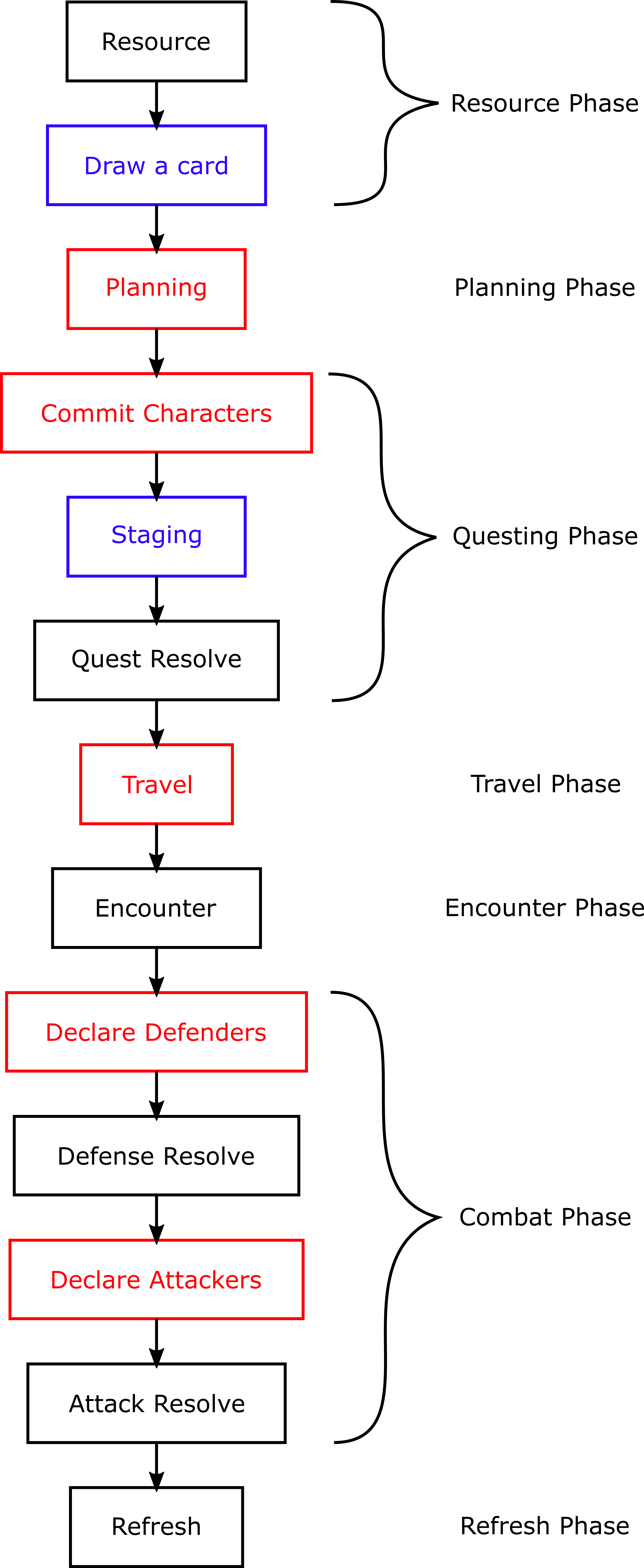}}
    \caption{Round structure: random (violet), decision (red), ruled (black) stages.}
    \label{fig:turn_structure}
\end{figure}

A diagram of one round of the game is shown in Fig.~\ref{fig:turn_structure}. There are 13 stages divided into seven round phases. Six of these stages are a simple implementation of the described game rules (marked in black), two of them contain unpredictable random actions (violet), while the other five are stages (red) in which the player must make a decision.

In LoTR, the player's task is to complete the scenario, consisting of three \textit{quest cards}. Each \textit{quest card} has a specific number of \textit{progress points} that must be obtained to complete a given stage of the scenario. The player receives \textit{progress points} by playing cards from his hand and then assigning them to the expedition. Opponents drawn randomly from \textit{encounter deck} hinder the progress of the expedition, additionally in the defensive phase they attack the player, dealing damage to hero and ally cards. In addition to opponents, heroes and allies, there are other types of cards in the game: places, events, and items. Places are destination cards, where player can travel to; events can be drawn from \textit{encounter deck}, they affect the player in a negative way; items are kind of attachments to heroes, giving them buffs.




\section{Agent players}
As represented on Fig.~\ref{fig:turn_structure}, there are 5 decision stages within the game: \textit{Planning}, \textit{Commitment}, \textit{Travel} and finally \textit{Declaring Defenders} and \textit{Attackers}. The stages of \textit{Travel} and \textit{Declaring Attackers} are simple for expert players, so they are not considered in subsequent analyzes. The other stages can be resolved in different ways, one of them may be random. Decisions can also be carried out by a reduced rule-based player with a simple logic implemented~\cite{cowling2012}. So finally, the analysis features 4 types of player agent: random, rule-based (hereafter called expert) and two MCTS versions: flat and full-UCB.

Stages from Fig.~\ref{fig:turn_structure} come sequentially one after another and action possibilities in a game node strictly depend on what decision has been taken in the antecedent. In \textit{Commitment} stage player submits characters to the quest, he can select one or more heroes or characters already purchased during the \textit{Planning}. Typically player purchase 1-2 cards from his hand, so for commitment he forms subset from a group of 5 cards. Size of the subset depends on current summary threat of cards in \textit{staging area} - once can be zero, other round 3 for example. The number of enemies in \textit{encounter area} determine the quantity of declared defenders - for instance if there are 3 enemies, player have to assign 3 of 5 characters, so it yields 10 subsets. To sum up - the action space size varies unpredictably depending on what happened in the past.

\subsection{Agent 1. Random choice}

\textit{Planning} stage consists of random selection of cards in hand and then checking out if they are possible to play in according to the game rules.

\textit{Commit Characters}, by this stage agent draws a subset of characters in play, if their total \textit{willpower} is higher than the total \textit{threat} of the cards in \textit{staging area}: commits it the quest, otherwise reject the subset and draw again.

In \textit{Declare Defenders} stage agent samples one character for every enemy in \textit{engagement area}. 

\subsection{Agent 2. Expert decision}

In \textit{Planning} cards from the \textit{Spirit} sphere are welcomed, because of their \textit{willpower} stats, which plays crucial role in the next decision stage. Likewise it is worth to play in \textit{Gandalf}, but his cost of 4, makes him affordable for the agent after a couple of rounds. In early game agent deals with few tokens in his \textit{resource pool}, therefore cheap cards like \textit{Wandering Took}, \textit{Gondorian Spearman} or \textit{Veteran Axehand} seem to be an adequate choice.

During \textit{Questing} \textit{Gandalf} and the \textit{Spirit} sphere characters come to play in a following manner - they get committed as long as total \textit{willpower} exceeds total \textit{threat} of the cards in \textit{staging area}. This condition increase the probability of success in quest resolution, which approach the agent to win the whole game.

In \textit{Defense} agent attempts at first declare allies as defenders, due to their lower opportunity cost in case of death unlike the hero cards. It is crucial to the gameplay keeping agent's heroes safe and sound, because they provide resource tokens, every removal of a hero gravely affects future rounds - the number of \textit{resource pools} gets reduced immediately.

\subsection{Agent 3. Flat Monte Carlo}

The idea behind flat Monte-Carlo tree search, is to create only first layer of tree expansion. 
\begin{enumerate}
    \item \textit{Expansion}
    \item \textit{Simulation}
    \item \textit{Selection}
\end{enumerate}

\textit{Expansion} is identical to full UCB method, the specified number of playouts (\textit{playoutBudget}) is divided into all children and \textit{Simulation} is performed. In \textit{Selection} the child with the highest number of wins is selected.

\subsection{Agent 4. MCTS-UCB}

Monte Carlo Tree Search is an algorithm for taking optimal decisions through sequentially built trees based on random sampling. The main advantage of this method is a utility function of a leaf in a simple form, which allows evaluation of a single step in decision process. The function called \textit{Upper Confidence Bound}:

\begin{equation}
UCB_{j} = x_{j} + C \sqrt{\frac{2 \ln n}{n_{j}}}
\end{equation}
where $x_j$ - average number of playouts won starting from given game state $j$, $n$ - the number of visits in parent of node $j$, $n_j$ - the number of node $j$ visits, $C$ - arbitrary chosen constant between 0 and 1. MCTS consists of 4 steps repeated until a certain playout budget is reached:
\begin{enumerate}
    \item \textit{Selection}
    \item \textit{Expansion}
    \item \textit{Simulation}
    \item \textit{Backpropagation}
\end{enumerate}

In the selection phase, \textit{UCB} is calculated for all leaves from the given parent, the node with the highest value is selected as the new parent and the selection process is repeated until final node is reached. In \textit{Expansion}, new leaves are added to leaf node, of course within the rules of the game - \textit{legal moves}. In the third phase, playouts to the terminal state of the game are performed with leaf node. During \textit{Backpropagation} statistics of the number of won playlists and the number of visits for all nodes up to root are updated.

The task of this agent is in accordance with the philosophy of \textit{Monte Carlo Tree Search} is to choose the most favorable move in a given phase of the game, i.e. the node with the highest utility function. Due to the fact that the tree is only developed to the first level, a simplified version of the utility function was used in the form of \textit{score}. The nodes from which the playouts are played must be created according to the circumstances of the given phase of the game, \textit{findLegals} functions have been implemented for each required decision. In \textit{Planning}, legal moves are determined by checking that the player's resources allow you to buy the card if you create the node. For \textit{Questing} all combinations of available characters are considered, if the total \textit{willpower} of a given subset is greater than the total \textit{threat} cards in \textit{staging area}, then a node is created. In addition, a restriction was imposed on the rejection of subsets containing cards with \textit{willpower} equal to zero.

Combination subsets of cards are also created in \textit{Defense}, but with a strictly defined number of elements, corresponding to the number of opponents in \textit{engagement area}. These subsets can contain heroes already embedded in previous stages of the game, then at \textit{resolving} an unprotected attack occurs. To do this, \textit{legalNodes} drop subsets with their already dated allies - they can no longer be used for defense.

\textit{LegalNodes} for \textit{Strike back}, similar to the previous phase, are subsets with a limited number of elements resulting from the number of untapped cards and enemies in \textit{engagement area}. After specifying \textit{legalNodes}, deep copy of the game are created from each node, from which playouts are performed. The resulting \textit{score} is then saved to the node.

\section{Experiments}

Series of 1000 simulations had been preformed to analyze statistical properties of the agents. The simulations were run in parallel on CPU - host machine with 12-cores Intel i9-9920X processor and 128GB RAM. Spawning processes across the CPU were performed with python \textit{multiprocessing} package. After pooling the results of the simulations, post-processing was applied - for every experiment winrate with confidence interval was calculated. Binomial proportion confidence interval for 95\% confidence level is described by the equation:
\begin{equation}
    z \sqrt{\frac{p (1 - p)}{n}},
\end{equation}
where $z$ - 1.96 for 95\% confidence level, $p$ - winrate probability $n_s / n$, $n$ - total number of trials, $n_s$ - number of wins.

%


\begin{figure}[htbp]
    \centerline{\includegraphics[width=0.5\textwidth]{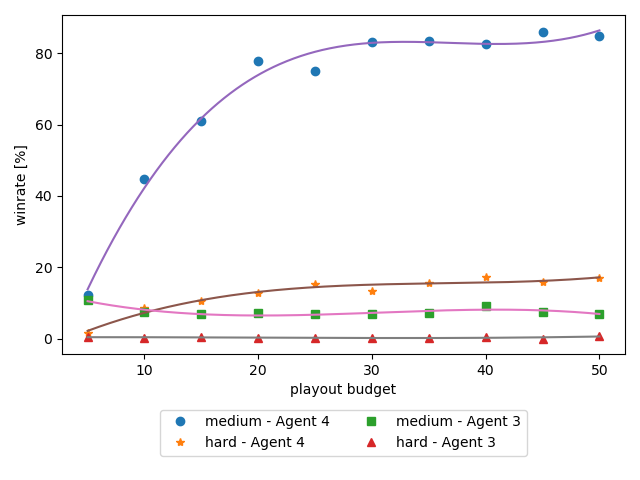}}
    \caption{Probability of winning (winrate) as a function of the number of playouts. The size of statistical sample is 1000 games. Agent\_3 and Agent\_4 are compared on two different complexity levels. Confidence intervals are not shown for more clear presentation.}
    \label{fig:pb}
\end{figure}

First of all, the impact of the playout budget was under investigation.  As depicted in Fig.~\ref{fig:pb}, Agent 4 perform a way better than his simplified brother Agent 3. Moreover this method turns out to be more vulnerable to the difficulty level - over 80\% for medium versus less than 20\% for hard level. It is also worth to note, that Agent 3 were not able to win at the top difficulty level regardless playout budget. The shape of the curve with saturation clearly suggests, that increasing the playout budget over 40 is redundant, therefore this value will be used in further considerations.

\begin{figure}[htbp]
    \centerline{\includegraphics[width=0.5\textwidth]{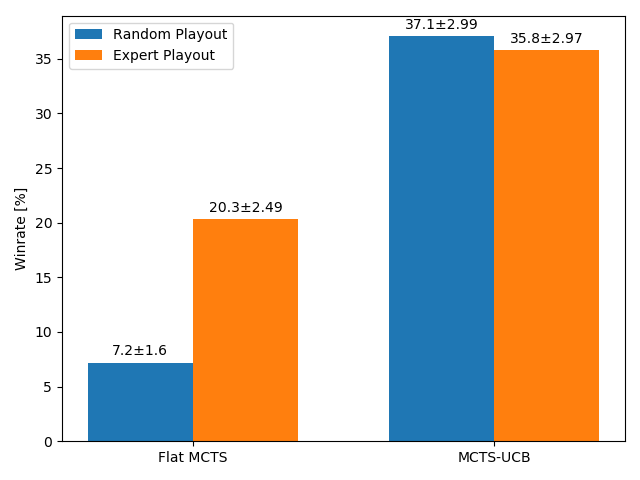}}
    \caption{Winrate comparison for with different types of playouts (random vs. expert). Playout budget is set to 40 and complexity level to medium.}
    \label{fig:playoutType}
\end{figure}

The second point of research was to determine volatility due to the type of playout, as seen in Fig.~\ref{fig:playoutType}. Results for  Agent 4 comes out quite equally - winrate about 36\%, what means that type of playout is not so important. 
However for Agent 3 it was observed that winrate in case of expert playouts are nearly three time higher. This could be treated as a confirmation that incorporation of the domain knowledge is significantly increasing performance of the agent.  

\begin{figure}[htbp]
    \centerline{\includegraphics[width=0.5\textwidth]{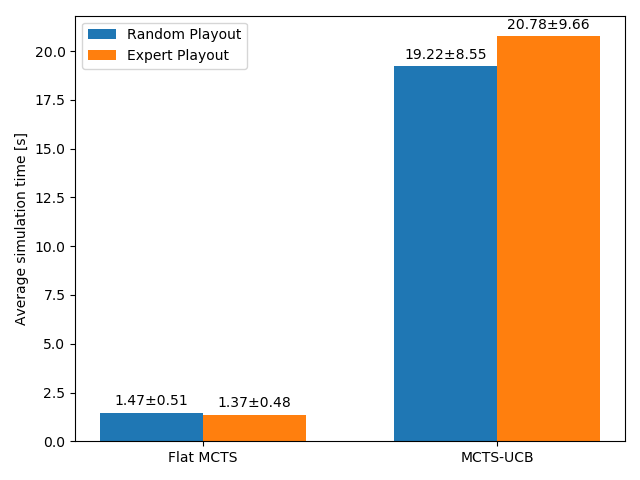}}
    \caption{Average simulation time for Agent 3 and Agent 4 for different types of playouts. Playout budget is set to 40 and complexity level to medium.}
    \label{fig:playoutTime}
\end{figure}

Type of the playout does not have important impact on the computation cost. Fig.~\ref{fig:playoutTime} shows that total simulation time in both configurations are nearly the same. So choosing playout strategy with expert knowledge with higher winrate Fig.~\ref{fig:playoutType}, does not deteriorate performance of the solver.

These two points lead to conclusion, that the optimal setup for playout strategy is: 40 repetitions in budget at expert mode. Within these circumstances the methods reach sufficient winrate at acceptable simulation time. 

\begin{figure}[htbp]
    \centerline{\includegraphics[width=0.52\textwidth]{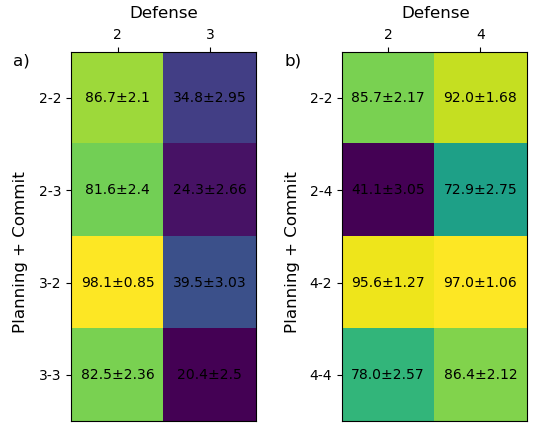}}
    \caption{Winrate for combination of agents on three different decision stages (Planning, Commit, Defence) (complexity medium); a) Agent 2 and Agent 3, b) Agent 2 and Agent 4}
    \label{fig:combinations}
\end{figure}

The next investigated problem is whether there is a type of the agent that has advantage over others is each of three decision stages.

Combinations of all agents were also broadly examined as seen in Fig.~\ref{fig:combinations}. Binomial proportion confidence interval theory has been used to estimate uncertainty of the probability of winning rate. These heatmaps suggest a vast influence of Agent 2 at \textit{Defense} phase in contrast to Agent 3, nothing similar happens to agent 4. More in detail, combinations including agent 2 at \textit{Questing} outperforms other arrangements. Final optimal choice is combination agents: 4 - 2 - 4 for which probability of win is over 97\%.

\section{Conclusions}

The Lord of the Rings is a popular cooperative card game. It should be classified as a multistage game with a high level of randomness, what creates serious challenge for computational intelligence methods. The MCTS algorithms are universal and powerful tool, however with high computational demands. Numerical experiments described in the paper have shown that incorporating expert knowledge significantly improves the performance of methods.
The second important conclusion is that the use of different methods at different stages of a multistage game allows to increase total winning rate.


\begin{thebibliography}{00}
\bibitem{b1} Magic: The Gathering, [online]https://magic.wizards.com/en
\bibitem{b2} Browne, A survey of monte carlo tree search methods, IEEE Transactions on Computational Intelligence and AI in games, vol. 4.,  pp. 1--43, 2012.
\bibitem{b3} http://hallofbeorn.com/LotR/Scenarios/Passage-Through-Mirkwood
\bibitem{cowling2012} Cowling, Peter I., Colin D. Ward, and Edward J. Powley. "Ensemble determinization in monte carlo tree search for the imperfect information card game magic: The gathering." IEEE Transactions on Computational Intelligence and AI in Games 4.4 (2012): 241-257.
\bibitem{b5} Bjarnason, Ronald, Alan Fern, and Prasad Tadepalli. "Lower bounding Klondike solitaire with Monte-Carlo planning." Nineteenth International Conference on Automated Planning and Scheduling. 2009.
\bibitem{b6} Chaslot, Guillaume MJ-B., Mark HM Winands, and H. Jaap van Den Herik. "Parallel monte-carlo tree search." International Conference on Computers and Games. Springer, Berlin, Heidelberg, 2008.
\bibitem{b7} Świechowski, Maciej, Tomasz Tajmajer, and Andrzej Janusz. "Improving hearthstone ai by combining mcts and supervised learning algorithms." 2018 IEEE Conference on Computational Intelligence and Games (CIG). IEEE, 2018.
\bibitem{b8} Fern, Alan and Paul Lewis. "Ensemble monte-carlo planning: An empirical study." Twenty-First International Conference on Automated Planning and Scheduling. 2011.
\bibitem{Turkay2012} Turkay Selen , Adinolf Sonam , Tirthali Devayani. "Collectible Card Games as Learning Tools", Procedia - Social and Behavioral Sciences. 46. 3701-3705. 2012,  10.1016/j.sbspro.2012.06.130
\bibitem{Ward2009} Ward, Colin D., and Peter I. Cowling. "Monte Carlo search applied to card selection in Magic: The Gathering." 2009 IEEE Symposium on Computational Intelligence and Games. IEEE, 2009.


\bibitem{Powley2012} E. J. Powley, D. Whitehouse and P. I. Cowling, "Monte Carlo Tree Search with macro-actions and heuristic route planning for the Physical Travelling Salesman Problem," 2012 IEEE Conference on Computational Intelligence and Games (CIG), Granada, 2012, pp. 234-241, doi: 10.1109/CIG.2012.6374161.

\bibitem{Segler2018} Segler, Marwin HS, Mike Preuss, and Mark P. Waller. "Planning chemical syntheses with deep neural networks and symbolic AI." Nature 555.7698 (2018): 604-610.


\end{thebibliography}
\end{document}